\documentclass{article}
\usepackage{ijcai20}

\usepackage{times}

\usepackage{soul}
\usepackage{url}
\usepackage[hidelinks]{hyperref}
\usepackage[utf8]{inputenc}
\usepackage[small]{caption}
\usepackage{graphicx}
\usepackage{amsmath}
\usepackage{booktabs}
\usepackage{booktabs}
\usepackage{algorithm}
\usepackage{algpseudocode}
\usepackage{subcaption}
\usepackage{float}
\title{Hierarchical Clustering using Auto-encoded Compact Representation for Time-series Analysis}
\author{
Soma Bandyopadhyay\footnote{Contact Author}\and
Anish Datta\And
Arpan Pal\\
\affiliations
TCS Research, TATA Consultancy Services ,Kolkata, India \\
\emails
\{soma.bandyopadhyay, anish.datta, arpan.pal\}@tcs.com
}

\begin{document}

\maketitle

\begin{abstract}
Getting a robust time-series clustering with best choice of distance measure and appropriate representation is always a challenge. We propose a novel mechanism to identify the clusters combining learned compact representation of time-series, Auto Encoded Compact Sequence (AECS) and hierarchical clustering approach. Proposed algorithm aims to address the large computing time issue of hierarchical clustering as learned latent representation AECS has a length much less than the original length of time-series and at the same time want to enhance its performance.Our algorithm exploits Recurrent Neural Network (RNN) based under complete Sequence to Sequence(seq2seq) auto-encoder and agglomerative hierarchical clustering with a choice of best distance measure to recommend the best clustering. Our scheme selects the best distance measure and corresponding clustering for both univariate and multivariate time-series. We have experimented with real-world time-series from UCR and UCI archive  taken from diverse application domains like health, smart-city, manufacturing etc. Experimental results show that proposed method not only produce close to benchmark results but also in some cases outperform the benchmark.
\end{abstract}

\section{Introduction}

Time-series learning quite often faces paucity of ground truth. Hence finding patterns, groups and sub groups become a challenge. Specifically dearth of label and cost of domain expert's knowledge to analyse time-series in healthcare, manufacturing, smart-city trigger the need of unsupervised learning approaches inevitable. 

In this work we propose a novel mechanism to identify the clusters combining learned compact representation of time-series along with hierarchical clustering approach \cite{friedman2001elements}. The ability of Hierarchical clustering to form non-convex clusters without requiring the number of clusters to be formed paves its way for many real world tasks. 
Our prime focus is to validate the proposed method across different IoT (Internet-of-Things) applications. 
We have performed experimental analysis on diverse applications like healthcare- ECG200, ECGFiveDays etc., machine and manufacturing- FordB (engine noise data), Wafer (semi-conductor fabrication) etc. and CricketX (human motion) etc. for smart-city. We have demonstrated our experimental analysis using both univariate and multivariate time-series from UCR Time Series Classification Archive\cite{bagnall2018uea} and UCI Machine Learning Repository\cite{Dua2019}. 

Our prime contributions are as follows: \\
\textbf{1. AECS and clustering on compact representation}: Auto encoded compact representation using Seq2Seq auto-encoder (AE)\cite{sutskever2014sequence} to generate the latent representation of time-series. It has under complete architecture, and the learned latent representation has a length much less than the original length of time-series. \\
Agglomerative hierarchical clustering is applied on AECS.\\
\textbf{2. Application of appropriate distance measure}: Chebyshev(CH), Manhattan(MA) and Mahalanobis(ML) are used as distance measures. These distance measures except ML take very low computing time to form the clusters.\\
\textbf{3. Choice of best distance measure}: We use Modified Hubert statistic($\mathcal{T}$) \cite{hubert1985} as an internal clustering validation measure. It computes the disagreement between every pair of time-series and evaluates separation between the clusters. We select the best clustering which has the highest $\mathcal{T}$. Here usage of Mahalanobis(ML) distance to evaluate $\mathcal{T}$ is our unique contribution. \\
\textbf{4. Extensive analysis on diverse time-series}: We study diverse and wide time-series from UCR Time Series Classification Archive \cite{bagnall2018uea} and UCI Machine Learning Repository\cite{Dua2019} spread across various application domains. 52 (41 univariate and 11 multivariate) time-series are considered for experiment and analysis of our proposed method. We use Rand Index (RI) \cite{rand1971objective} as external validation measure to compare our algorithm with other benchmark  algorithms. We demonstrate our proposed method is not only comparable to the established State-of-the-Art clustering algorithms like K-shape \cite{paparrizos2015k}, and hierarchical clustering with parametric derivative dynamic time warping (HC-DDTW)\cite{luczak2016} but outperforms them significantly.

Our algorithm is applicable both on univariate as well as multivariate time-series whereas the benchmark clustering methods are applicable only univariate time-series. In case of multivariate time-series we have compared the benchmarks from \cite{bagnall2018} and MLSTM-FCN\cite{karim2019}. Additionally we show that our mechanism is computationally efficient due to compact latent representation and choice of appropriate distance measure. This is very useful for real-world time-series having large sequence length.

\section{Related Works}

 
Hierarchical clustering(HC) algorithms create a hierarchy of clusters either using an bottom-up (Agglomerative) or top-down approach (Divisive) \cite{kaufman2009finding}. In agglomerative approach, two of the existing clusters are merged recursively while in divisive method, existing clusters are split into two. Hierarchical clustering does not require the prior information of number of clusters to be formed. It makes it ideally suitable for a number of real world problems where number of clusters to be formed is not available. Additionally its ability to form any arbitrary shaped and sized clusters makes it useful for many diverse applications. But the high computation cost of Hierarchical clustering acts as its major disadvantage.
Maciej Luczak proposed an algorithm \cite{luczak2016} based on agglomerative hierarchical clustering of time-series with a parametric Dynamic Time Warping (DTW) derivative (DDTW) distance measure showing HC outperforming K-Means. Though this approach addresses the challenge of finding a suitable distance measure, its drawback is longer computing time as finding DTW \cite{sakoe1978dynamic} between two time-series is extremely computation heavy.
Online Divisive-Agglomerative Clustering(ODAC)\cite{rodrigues2006} is a well known algorithm for hierarchical clustering of time-series which uses a top-down (Divisive) approach. Initially it assumes all the input time-series as a single cluster and then starts splitting it in different clusters in subsequent iterations. The splitting is continued until a certain heuristic condition is satisfied.

Among partitional based clustering, K-means \cite{macqueen1967some} algorithm is most popular one. 
K-Shape \cite{paparrizos2015k} is another partitional clustering algorithm applied vastly on time-series. It uses shape-based distance metric (SBD) to find similarity between two time-series. Using this SBD, centroids are computed using a shape-extraction method for each cluster and then assign points to the clusters whose centroids are nearest to them. Although partitional clustering algorithms takes low time, they suffer from a potential shortcoming of generating hyper-spherical clusters which may not fit in most of real world datasets. 

\section{Proposed method}

\begin{figure}[htbp]
\centerline{\includegraphics[width=90mm]{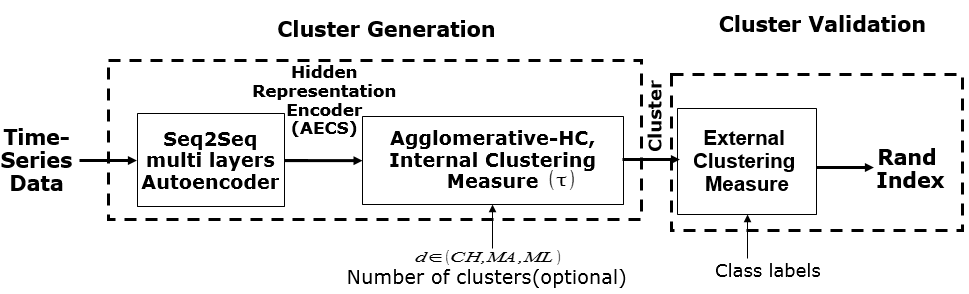}}
\caption{Functional Block of HC-AECS}
\label{fig}
\end{figure}

In this work we propose a novel method to perform clustering by learning Auto Encoded Compact Sequence (AECS), a latent representation and applying hierarchical clustering on AECS. Our method encompasses exploiting distance measures like ML (Mahalanobis)\cite{de2000mahalanobis} and low computing measures like CH (Chebyshev), MA (Manhattan)\cite{wang2013experimental} to perform agglomerative hierarchical clustering and selecting best choice of clusters applying Modified Hubert Statistic ($\mathcal{T}$) as internal clustering measure. The general functional block of our proposed approach is illustrated in Figure \ref{fig}. 
\vspace{-.2cm}
\subsection{AECS: Auto-encoded Compact Sequence }
\begin{figure}[htbp]
\centerline{\includegraphics[width=82mm]{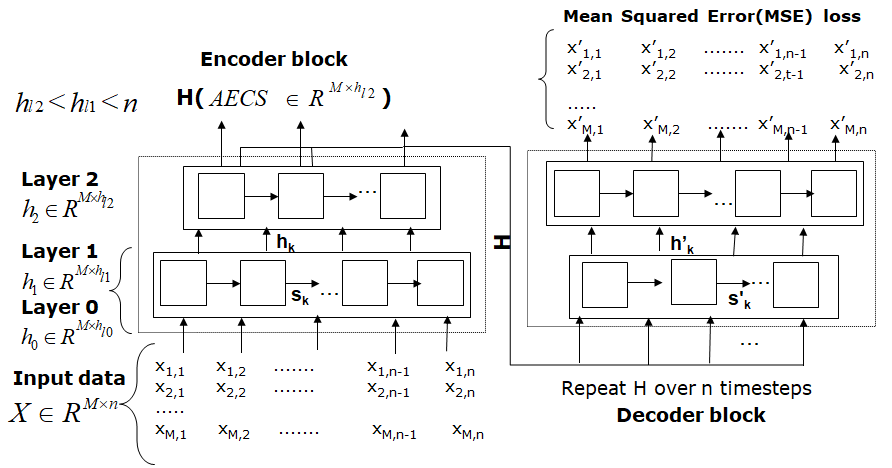}}
\caption{Schematic illustration of AECS using multilayer auto-encoder}
\label{fig1}
\end{figure}

In our method we first learn a compact representation of the time-series by using an Seq2Seq LSTM \cite{hochreiter1997long} auto-encoder. Let $X=\{X_1,X_2, .... ,X_M\}$ be a set of $M$ time-series, where $X_i=\{x_{i,1},x_{i,2},..,x_{i,k},..x_{i,n}\}$, $x_{i,k}$ gives the value of $i^{th}$ time-series's $k^{th}$ timestep, and n is the total number of timesteps or length of the time-series. In case of univariate, $x_{i}$ is scalar and is vector in case of multivariate time-series. We learn compact representation of time-series using multi layer Seq2Seq LSTM auto-encoder. The encoder block of the auto-encoder consists of an input layer $h_0$ having same number of units as number of timesteps and two hidden layers $h_1$ and $h_2$ having lengths $h_{l1}$ and $h_{l2}$ respectively. We propose Auto-Encoded Compact Sequence $AECS \in R ^{M \times h_{l2}}$,  which is the latent representation of layer $h_2$ of encoder. The length of AECS is $(h_{l2})$ where $h_{l2} < h_{l1} <n $. A schematic illustration of the auto-encoder model is presented in Figure \ref{fig1}.

We aim to learn a compact representation having a length much less than the original time-series to reduce the computing time as well as to capture important characteristics of time-series. AECS is an undercomplete representation hence bound to capture the important features of the input time-series.\\
Encoder: \\ $H_i= f(X_i)\gets h_{i,k},s_{i,k}\gets f_{LSTM}(h_{i,k-1},s_{i,k-1}x_{i,k})$; Decoder: \\ $X'_i = g(H_i)\gets h'_{i,k},s'_{i,k}\gets f_{LSTM}(h'_{i,k-1},s'_{i,k-1},h_{i,k})$ ..... (1), where $X \in R^{M \times n},H \in R^{M \times l_{ev}}$ and $X' \in R^{M \times n}$; $l_{ev}$: length of the encoder vector.
\\ Here $l_{ev} = length(AECS) = h_{l2}$. The reconstruction loss is Mean Square Error(mse): $l_{recon}=\dfrac{1}{Mn}\sum\limits_{i=1}^M \sum\limits_{j=1}^n(x_{i,j} -x'_{i,j})^2$ , where $X_i'=\{x'_{i,1},x'_{i,2},.....,x'_{i,n}\}$ is the reconstructed output for $i^{th}$ instance. 

\vspace{-.2cm}

\subsection{HC using diverse distance measure}
Hierarchical clustering (HC) produces a hierarchy of clusters formed either by merging or dividing existing clusters recursively. The iterative process of agglomeration/division of the clusters can be visualised in form of a tree like structure known as dendrogram. Each node of the tree/dendrogram contains one or more time-series ($X_i$) representing a cluster. Here we use agglomerative hierarchical clustering which takes each input time-series as an individual cluster and then start successively merging pair of clusters having highest similarity between them until a single cluster is formed. Hence it measures similarities among the pair of time-series to put them into same cluster and the dissimilarities between the pair of groups of time-series to take the decision of further fusing the clusters. We have used average linkage \cite{friedman2001elements} to obtain pairwise inter-cluster dissimilarities. In average linkage, distance $a_d$ between clusters $C_0$ and $C_1$ is : $a_d(C_0,C_1)= \dfrac{1}{|C_0| \cdot |C_1|} \sum\limits_{X_i \in C_0} \sum\limits_{X_j \in C_1} dist(X_i,X_j)$. ..... (2) \\
$|C_i|$ denotes cardinality or number of members in cluster i. Average linkage aims to form compact and relatively far apart clusters than other linkage methods like single or complete by keeping average pairwise dissimilarity. Usage of linkage methods like single or complete are scope of future study.

In this work we have used following three different distance measures to find the dissimilarity between two time-series (say $X_i$ and $X_j$, where $X_i=\{x_{i,1},x_{i,2},..,x_{i,k},..x_{i,n}\}$ and $X_j=\{x_{j,1},x_{j,2},..,x_{j,k},..x_{j,n}\}$, $x_{i,k}$ and $x_{j,k}$ indicates $k^{th}$ timestep of $i^{th}$ and $j^{th}$ time-series respectively) as well as to obtain average linkage among the clusters as follows. \\
\textbf{Chebyshev Distance}: $max_{k}(|x_{i,k}-x_{j,k}|)$ .....(3): The maximum distance between two data points in any single dimension.\\
\textbf{Manhattan Distance}: $\sum\limits_{k=1}^T{|x_{i,k}-x_{j,k}|}$.....(4): This belongs to Minkowski family which computes distance travelled to get from one data point to the other if a grid-like path is followed.\\
\textbf{Mahalanobis Distance}: $\sqrt{(X_i-X_j)^T \cdot C^{-1} \cdot {(X_i-X_j)}}$...(5): Finds the distance between two data points in multidimensional space. Here $X_i$ and $X_j$ are two time-series and $C$ is the co-variance matrix between $X_i$ and $X_j$.

\subsection{Choice of best clustering and corresponding distance measure}
We use Modified Hubert statistic ($\mathcal{T}$), an internal clustering measure which evaluates sum of distance between each pair of instances weighted by distance between the centers of clusters which they belong to. We rank clustering formed by different distance measures based on $\mathcal{T}$. Best clustering has highest $\mathcal{T}$. In case two or more distance measures produces the highest value of $\mathcal{T}$, we report all of them as the best measures. We have measured disagreements between pairs of time-series and separation between clusters using ML (Mahalanobis) to evaluate $\mathcal{T}$, which is one of our prime contributions. 
\\
$\mathcal{T} = \frac{2}{n(n-1)} \sum\limits_{X_i \in X} \sum\limits_{X_j \in X} d(X_i,X_j) d(c_i,c_j)$ .....(6) \\
$d(X_i,X_j)= d_{ML}(X_i,X_j)$ ; $d(c_i,c_j) = d_{ML}(c_i,c_j)$.....(7)
where $C_i$ represents $i^{th}$ cluster and $c_i$ is the center of cluster $C_{i}$, $d_{ML}(X_i,X_j)$ is  Mahalanobis distance between time-series $X_i$ and $X_j$ and $d_{ML}(c_i,c_j)$ is Mahalanobis distance between centers of clusters to which $X_i$ and $X_j$ belongs.
\vspace{-.465 cm}
\begin{algorithm}[H]
\caption{HC-AECS: Hierarchical clustering with auto-encoded compact latent representation}
\label{alg:algorithm}
\textbf{Input}: Time-series: $X \in R^{M \times n}$, Number of time-series: $M$, Sequence length: $n$, Encoder hidden layers length: $h_{l1}, h_{l2}$ ; $h_{l2} < h_{l1} <n$, Epochs: $e$, Batch size: $b$, Learning rate: $lr$, Number of clusters: $K$ (optional)\\
\textbf{Output}:  Best clustering with corresponding distance measure
\begin{algorithmic}[1] 
\Procedure{AECS}{$X,h_{l1},h_{l2},e,b,lr$}
\State Create a multilayer auto-encoder using Eq. (1).
\State Get latent representation from layer $h_2$ of encoder.

$AECS \gets H_{lstm} ( X,h_{l1},h_{l2},e,b,lr,SGD)$
\State \textbf{return} $AECS \in R ^{M \times h_{l2}}$ 
\EndProcedure

\Procedure{Cluster\_AECS}{AECS,$D$=\{CH,MA,ML\}}
\For{distance metric $d_i$ in $D$}
\State  $C \gets HC(AECS,K,d_i,linkage=average)$ 
\State $Clusters[d_i] \gets C$  
\EndFor
\State \textbf{return} $Clusters$
\EndProcedure

\Procedure{BestCluster}{Clusters,$D$=\{CH,MA,ML\}}
\State $max_{\mathcal{T}} \gets 0$
\State $d_{best} \gets null$
\For{distance metric $d_i$ in $D$}
\State $C \gets Clusters[d_i]$
\State $\mathcal{T} \gets Modified\_Hubert\_Statistic(C)$ ,based on Eq.(6) and (7)
\If {$max_{\mathcal{T}} < \mathcal{T}$}
\State $d_{best} \gets \O$
\State $max_{\mathcal{T}} \gets \mathcal{T}$
\State $d_{best} \gets d_i$
\ElsIf {$max_{\mathcal{T}} = \mathcal{T}$} 
\State  $d_{best} \gets d_{best} \cup d_i$
\EndIf
\EndFor
\State \textbf{return} $\{Clusters[d_{best_i}],d_{best_i}\}  , \forall i$  
\EndProcedure
\end{algorithmic}
\end{algorithm}

\subsection{Validation}
We compute Rand Index (RI) \cite{rand1971objective}, one of the external clustering validation measures to compare the performance of our proposed mechanism with the benchmark results. RI is a cluster validation measure which considers all pairs of time-series and counts number of pairs assigned in same or different clusters formed by our algorithm to the true clusters based on the given labels. 
$RI : \frac{TP+TN}{TP+FP+FN+TN}$ .....(8) , where the symbols denote cardinalities of sets of pairs: TP (true positive) denotes elements which are assigned to same cluster that belongs to the same class; TN (true negative) denotes elements which are assigned to different clusters that belongs to different classes; FP (false positive) denotes elements which are assigned to different clusters that belongs to the same class; FN (false negative) denotes elements which are assigned to same cluster that belongs to different classes. 

\section{Experimental Analysis}
\subsection{Data Description}
We use univariate and multivariate time-series from UCR Time Series Classification Archive \cite{bagnall2018uea} to evaluate the performance of our proposed algorithm. Additionally we perform our analysis on 2 multivariate datasets from UCI Machine Learning Repository \cite{Dua2019}. Description of univariate and multivariate time-series are presented in Tables \ref{uni_dd}  and \ref{multi_dd} respectively. Each dataset has default train-test split. In our approach we merge the train and test sets and perform Z-normalisation on the merged data.
\vspace{-.4cm}
\begin{table}[ht]
\centering
\caption{Dataset Description of UCR univariate time-series}
\label{uni_dd}
\resizebox{0.75\columnwidth}{!}{\begin{tabular}{lrrrr}
\toprule
\textbf{Dataset} & \textbf{\#Train} & \textbf{\#Test} & \textbf{Length} & \textbf{\#class} \\ 
\midrule
DistalPhalanxOAG & 400 & 139 & 80 & 3 \\ 
DistalPhalanxOC & 	600	& 276 &	80 & 2 \\ 
DistalPhalanxTW & 400 & 139 & 80 & 6 \\ 
MiddlePhalanxOAG & 400 & 154 & 80 & 3 \\ 
MiddlePhalanxOC	& 600 &	291 & 80 &	2 \\ 
MiddlePhalanxTW & 399 & 154 & 80 & 6 \\ 
ProximalPhalanxOC & 600 & 291 &	80 & 2 \\ 
ProximalPhalanxTW & 400 & 205 & 80 & 6 \\ 
Beef & 30 & 30 & 470 & 5 \\ 
Earthquakes & 322 & 139 & 512 &	2 \\
Coffee & 28 & 28 &	286 &	2  \\ 
Fish & 175 & 175 & 463 & 7 \\ 
Ham & 109 & 105 & 431 & 2 \\  
Strawberry & 613 & 370 & 235 & 2 \\ 
DiatomSizeReduction	& 16 & 	306 & 345 &	4 \\ 
Wine & 57 & 54 & 234 & 2  \\  
ChlorineConcentration & 467 & 3840 & 166 & 3 \\ 
SyntheticControl & 300 & 300 & 60 & 6 \\ 
TwoPatterns & 1000 & 4000 & 128 & 4 \\ 
CricketX & 390 & 390 & 300 & 12 \\ 
CricketY & 390 & 390 & 300 & 12 \\ 
CricketZ & 390 & 390 & 300 & 12 \\ 
ECG200 & 100 & 100 & 96 & 2 \\ 
ECG5000 & 500 & 4500 & 140 & 5 \\ 
ECGFiveDays & 23 & 861 & 136 & 2 \\ 
TwoLeadECG & 23 & 1139 & 82 & 2 \\ 
Lightning7 & 70 & 73 & 319 & 7 \\ 
Plane & 105 & 105 &	144	& 7 \\ 
SonyAIBORobotSurface1 &  20 & 601 &	70 & 2 \\ 
SonyAIBORobotSurface2 & 27	& 953 &	65 & 2 \\ 
ArrowHead & 	36 & 175 &	251 & 	3 \\ 
SwedishLeaf & 500 & 625 & 128 & 15 \\ 
Yoga & 	300	& 3000 & 426 & 2 \\ 
Car	& 60 &	60 &	577 & 	4 \\ 
GunPoint & 50 &	150 & 150 & 2 \\ 
Adiac & 390 & 391 & 176 & 37 \\ 
ToeSegmentation1 & 40 &	228 & 277 &	2 \\ 
ToeSegmentation2 &	36 & 130 &	343 & 2 \\ 
Wafer & 1000 & 6164 & 152 & 2 \\ 
FordB & 	3636 &	810 &	500 &	2 \\ 
BME & 	30 & 	150 & 	128 & 	3 \\ 
\bottomrule
\end{tabular}}
\end{table}

\begin{table}[hbt!]
\centering
\caption{Dataset Description of UCR and UCI Multivariate time-series}
\label{multi_dd}
\resizebox{0.9\columnwidth}{!}{\begin{tabular}{llrrrrr}
\toprule
\textbf{Archive} & \textbf{Dataset} & \textbf{\#Train} & \textbf{\#Test} & \textbf{Length} & \textbf{\#class} & \textbf{\#dim}\\ 
\midrule
& AtrialFibrilation & 15 & 15 & 640 & 3 & 2 \\
& ERing & 30  & 30 &	65 & 6 & 4 \\ 
& FingerMovements & 316 &	100 & 50 & 	2 & 28 \\
& HandMD & 160 & 74 & 400 & 4 & 10 \\
UCR & Handwriting & 150 & 850 & 152 & 26 & 3 \\ 
& Libras & 180 & 180 & 45 & 15 & 2 \\
& NATOPS & 180 & 	180 &  	51 & 	6 & 24 \\
& SelfRegulationSCP2 & 200 &	180 &	1152 & 2 & 7 \\
& StandWalkJump & 12 & 15	& 2500 & 3 & 4 \\\hline \\ 
UCI & Wafer & 298 &	896 & 104-198 & 2 & 6 \\ 
& GesturePhase & 198 & 198 & 4-214 & 5 & 18\\
\bottomrule
\end{tabular}}
\end{table}

\subsection{Latent representation} 
\label{subsec:l1}
We first compute Auto-Encoded Compact Representation for time-series following algorithm HC-AECS.AECS() stated in proposed method. We have used Mean-Squared Error (MSE) as loss function and Stochastic Gradient Descent (SGD) \cite{robbins1951stochastic} as optimizer with learning rate ($lr$) as 0.004 and  momentum as 0. We use batch size 32. 

Let us take a representative dataset Adiac to explain thoroughly. It is an univariate time-series with 781 instances (M  = 781) where each time-series is of length 176 (n=176) i.e $X \in R^{781 \times 176}$. The input layer $h_0$ produces a representation of the same length as that of the number of time steps i.e $h_{0} \in R^{781 \times 176}$. This representation is passed to first hidden layer $h_1$ which scales down the length of each instance to $h_{l1}=16$ ($h_{1} \in R^{781 \times 16}$). This compressed sequence is fed to layer $h_2$ of length $h_{l2}=12$ where we get a further compressed representation for each $X_i$ of Adiac as $AECS_{Adiac}\in R^{781 \times 12}$. 

Also let us consider a multivariate dataset ERing having 60 instances (M=60) of time-series length 65 (n=65) with 4 dimensions i.e $X \in R^{60 \times 65 \times 4}$. Each timestep consists of a vector of size 4 corresponding to the dimensions. Here after passing the dataset through the input layer $h_0$ we get a representation for each time-series of length equal to the number of timesteps. Correspondingly layers $h_1$ and $h_2$ produces representations of length $h_{l1}$ and $h_{l2}$ for each time-series respectively.

\subsection{Study on diverse distance measure and choice of best clustering}
\begin{figure}[htbp]
\centerline{\includegraphics[width=75mm]{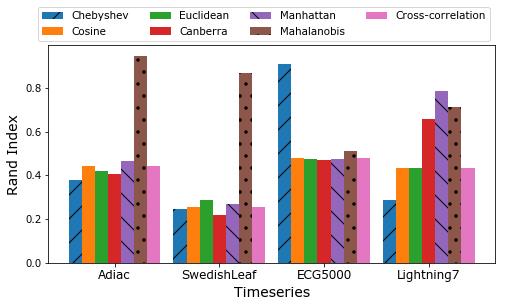}}
\caption{Comparison of RI using HC with 7 distance measures on 4 raw representative time-series. The measures marked with patterns are used by our algorithm}
\label{dist}
\end{figure}

We started our analysis using 7 different distance measures - Chebyshev, Cosine, Euclidean, Canberra, Manhattan, Mahalanobis and Cross-correlation. After extensive analysis, we have concluded that Chebyshev, Manhattan and Mahalanobis performs better than the other measures on raw time-series as well as on proposed compact representation of time-series - AECS. Figure \ref{dist} depicts RI measures of HC on raw time-series using the above mentioned seven distance measures on four representative datasets. We can see either Chebyshev (CH), Manhattan (MA), Mahalanobis (ML) performs best in each one of these datasets. Hence, we use Chebyshev, Manhattan and Mahalanobis as the distance measures in our algorithm.

Next we calculate $\mathcal{T}$ based on Eq.6 and Eq.7 and  HC-AECS.BestCluster() on the clusterings formed by above concluded three distance measures, $\mathcal{T}_{CH}$ for CH, $\mathcal{T}_{MA}$ for MA and $\mathcal{T}_{ML}$ for ML. Extending the example presented in section \ref{subsec:l1}, for Adiac we obtained $\mathcal{T}_{CH} = 1.38$, $\mathcal{T}_{MA} = 1.56$,  $\mathcal{T}_{ML} = 1.90$. Hence we choose clusters formed by ML as it has highest $\mathcal{T}$.  While validation we see RI of clusters formed by ML is indeed the highest among the three RI measures ( RI (CH) = 0.698, RI (MA) = 0.780, RI (ML) = 0.926). 

\begin{figure}[htbp]
\centerline{\includegraphics[width=80mm]{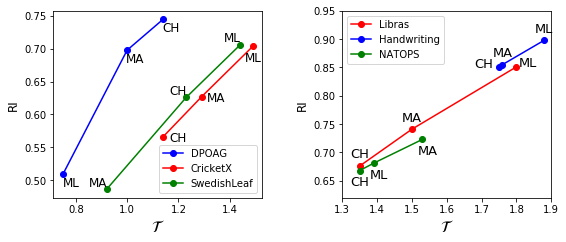}}
\caption{$\mathcal{T}$ vs RI for 3 distance measures CH, MA, ML for 3 univariate(left) and 3 multivariate(right) datasets, depict highest value of $\mathcal{T}$ has max RI }
\label{fig2}
\end{figure}

We illustrate further in Figure \ref{fig2} the plot of $\mathcal{T}$ vs RI for the said 3 distance measures on 3 univariate (DistalPhalanxOAG (DPOAG), CricketX, and SwedishLeaf) and 3 multivariate (Libras, Handwriting and NATOPS) time-series. In this figure we can observe that for each time-series, $\mathcal{T}$ and RI are linearly dependent signifying the clustering having the best $\mathcal{T}$ also has the highest RI using either CH, MA, ML. As depicted in the figure for the univariate time-series CH produces the best clustering for DPOAG and ML produces best clustering for CricketX and SwedishLeaf. Whereas for multivariate time-series, MA performs best for NATOPS and ML performs best for Libras and Handwriting.    

\begin{table*}[ht]
\centering
\caption{Rand Index (RI) comparison of HC-AECS with benchmark clustering algorithms on UCR time-series}
\label{hcdtw_10}
\resizebox{0.75\textwidth}{!}{\begin{tabular}{lrrrrrrrrrrr}
\toprule
 &  & & \multicolumn{3}{c}{\textbf{HC-L}}  & \multicolumn{6}{c}{\textbf{HC-AECS}}\\ \cline{4-6} \cline{7-12}
\textbf{Dataset} & \textbf{HC-DDTW}  & \textbf{K-Shape} &   &  &  &  \multicolumn{2}{c}{\textbf{CH}} & \multicolumn{2}{c}{\textbf{MA}} & \multicolumn{2}{c}{\textbf{ML}}\\ \cline{7-12}
& & & \textbf{CH} & \textbf{MA} & \textbf{ML}  & \textbf{RI} & \textbf{$\mathcal{T}$} & \textbf{RI} & \textbf{$\mathcal{T}$} & \textbf{RI} & \textbf{$\mathcal{T}$}\\  
\midrule
DistalPhalanxOAG & 0.709  & 0.598 & 0.706 & \textit{0.710} & 0.458 & \textbf{0.746} & \textit{1.15} & 0.694 & 1.00 & 0.466 & 0.23\\  
DistalPhalanxOC & \textbf{0.527} & 0.500 & \textbf{0.527} & \textit{0.527} & 0.502 & \textbf{0.527} & \textit{0.01} & \textit{0.527} & \textit{0.01} & 0.509 & 0.01  \\  
DistalPhalanxTW & 0.862 & 0.700  & \textit{0.814} & 0.812 & 0.593 & \textbf{0.885} & \textit{1.31} & 0.846 & 1.21 & 0.560 & 1.18\\  
MiddlePhalanxOAG & 0.729 & \textbf{0.733} & 0.390 & 0.387 & \textit{0.518}  & \textit{0.702} & \textit{1.08} & 0.688 & 0.98 & 0.509 & 0.96 \\  
MiddlePhalanxOC & 0.500 & 0.505 & \textbf{\textit{0.529}} & \textit{0.529} & 0.521  & 0.508 & 0.37 & \textit{0.525} & \textit{0.45} & 0.517 & 0.31 \\  
MiddlePhalanxTW & 0.802 & 0.719 & 0.713 & \textit{0.796} & 0.673  & \textbf{0.830} & \textit{1.48} & 0.795 & 1.35 & 0.502 & 0.96\\  
ProximalPhalanxOC & 0.535 & 0.533 & \textbf{0.565} & \textit{0.565} & 0.544 & 0.555 & 0.06 & \textit{0.564} & \textit{1.00} & 0.505 & 0.83\\  
ProximalPhalanxTW & \textbf{0.880} & 0.769 & 0.792 & \textit{0.794} & 0.596  & 0.822 & 1.18 & \textit{0.844} & \textit{1.31} & 0.594 & 1.05\\  
Beef & 0.582 & \textbf{0.710} & 0.638 & 0.373 & \textit{0.656}  & 0.373 & 0.49 & 0.373 & 0.49 & \textit{0.582} & \textit{1.22}\\  
Earthquakes & 0.541 & 0.640 & 0.674 &  0.672 & 0.520 & 0.672 & 0.01 & \textbf{0.680} & \textit{0.25} & 0.680 & 0.01 \\
Coffee & 0.491 & \textbf{0.584} & 0.494 & 0.494 & \textit{0.497}  & \textit{0.514} & \textit{0.95} & 0.492 &  0.60 & 0.501 & 0.33 \\   
Fish & 0.181 & \textbf{0.817} & 0.435 & 0.448 & \textit{0.740}  & 0.531 & 1.04 & \textit{0.591} & \textit{1.20} & 0.556 & 1.14 \\   
Ham & 0.498 & 0.498 & 0.498 & \textit{0.499} & 0.498   & 0.498 & 0.04 & 0.498 & 0.04 & \textbf{0.502} & \textit{0.31} \\   
Strawberry & 0.504 & 0.500 & 0.509 & 0.523 & \textit{0.540}   & 0.500 & 0.46 & 0.500 & 0.45 & \textbf{0.541} & \textit{1.00} \\  
DiatomSizeReduction & 0.296 & \textbf{0.908} & 0.296 & 0.306 & \textit{0.575} & 0.758 & 1.09 & \textit{0.763} & 0.99 & 0.604 & 0.97 \\  
Wine & 0.499 & 0.496 & \textit{0.499} & \textit{0.499} & 0.495  & 0.496 & 0.11 & \textbf{0.502} & \textit{0.91} & 0.496 & 0.04\\  
ChlorineConcentration & 0.403 & \textbf{0.533} & 0.403 & 0.413 & \textit{0.416}  & 0.507 & 1.02 & 0.493 & 0.88 & \textit{0.509} & \textit{1.10}\\  
SyntheticControl & 0.875 & \textbf{0.911} & \textit{0.834} & 0.692 & 0.687  & \textit{0.840} & \textit{1.49} & 0.839 & 1.48 & 0.502 & 0.97\\  
TwoPatterns & \textbf{0.848} & 0.691 & 0.255 & \textit{0.556} & 0.484  & 0.557 & 0.99 & \textit{0.635} & \textit{1.35} & 0.514 & 1.05\\  
CricketX & 0.777 & \textbf{0.862} & 0.123 & 0.709 & \textit{0.820}  & 0.548 & 1.10 & 0.551 & 1.11 & \textit{0.773} & \textit{1.65} \\  
CricketY & 0.688 & \textbf{0.862} & 0.114 & 0.748 & \textit{0.841}  & 0.557 & 1.09 & 0.547 & 1.07 & \textit{0.724} & \textit{1.53} \\  
CricketZ & 0.710 & \textbf{0.863} & 0.127 & 0.671 & \textit{0.843}  & 0.553 & 1.11 & 0.537 & 1.07 & \textit{0.799} & \textit{1.72} \\  
ECG200 & 0.537 & 0.546 & 0.531 & \textbf{\textit{0.555}} & 0.507  & 0.504 & 0.86 & 0.514 & 0.92 & \textit{0.546} & 0.26 \\  
ECGFiveDays & 0.499 & 0.516 & 0.499 & \textit{0.503} & \textit{0.503}  & \textbf{0.574} & 0.84 & 0.561 & 0.81 & 0.506   & 0.33\\  
ECG5000 & 0.891 & 0.667 & \textbf{0.910} & 0.474 & 0.509  &  0.744 & 1.33 & \textit{0.849} & \textit{1.58} & 0.528 & 1.07\\  
TwoLeadECG & 0.500 & \textbf{0.501} & \textit{0.500} & \textit{0.500} & \textit{0.500}  & \textbf{0.501} & \textit{0.17} & \textit{0.501} & \textit{0.17} & \textit{0.501} & 0.00 \\  
Lightning7 & 0.604 & 0.782 & 0.287 & \textbf{0.784} & 0.713  & \textit{0.627} & \textit{1.36} & 0.624 & 1.24 & 0.622 & 1.27\\  
Plane & \textbf{1.000} & 0.948 & \textit{0.902} & 0.838 & 0.735  & 0.690 & 1.21 & \textit{0.719} & \textit{1.29} & 0.667  & 1.29 \\  
SonyAIBORobotSurface1 & 0.499 & \textbf{0.659} & \textit{0.506} & \textit{0.506} & 0.500  & 0.504 & 0.09 & 0.502 & 0.10 & \textit{0.507} & \textit{0.14} \\  
SonyAIBORobotSurface2  & 0.534  & 0.558 & \textit{0.534} & 0.527 & 0.514  & 0.585 & 0.92 & \textbf{0.598} & \textit{0.93} & 0.526 & 0.01\\  
ArrowHead & 0.349 & \textbf{0.601} & 0.344 & 0.344 & \textit{0.533} & 0.341 & 0.06 & 0.343 & 0.08 & \textit{0.474} & \textit{0.80} \\  
SwedishLeaf & 0.348 & \textbf{0.925} & 0.245 & 0.269 & \textit{0.869}  & 0.400 & 0.73  & 0.422 & 0.78 & \textit{0.805}  & \textit{1.70}\\  
Yoga & \textbf{0.504} & 0.500 & 0.500 & 0.500 & \textit{0.502} & \textit{0.503} & 0.35 & \textit{0.503} & \textit{0.39} & 0.502 & 0.00\\  
Car & 0.498 & \textbf{0.699} & \textit{0.509} & 0.496 & 0.403 & \textit{0.608} & \textit{1.37} & 0.606 & 1.37 & 0.507 & 1.21\\  
GunPoint & 0.498 & 0.503 & 0.499 & \textbf{0.514} & 0.498   & \textit{0.498} & \textit{0.04} & \textit{0.498} & \textit{0.04} & \textit{0.498}   & \textit{0.04}\\  
Adiac & 0.683 & \textbf{0.955} & 0.377 & 0.466 & \textit{0.947}  & 0.698 & 1.38 & 0.780 & 1.56 & \textit{0.926} & \textit{1.90}\\  
ToeSegmentation1 & \textbf{0.505} & 0.498 & 0.499 & 0.498 & 0.498  & 0.499 & 0.94 & \textit{0.501} & \textit{1.00} & 0.499   & 1.00 \\  
ToeSegmentation2 & \textbf{0.665} & 0.608 & \textit{0.626} & 0.497 & 0.602  & 0.501 & 0.01 & 0.521 & 0.97 & \textit{0.614}   & \textit{1.02} \\  
Wafer & 0.534 & 0.527 & \textit{0.808} & 0.534 & 0.770  & 0.534 & 0.89 & 0.534 & 0.89 & \textbf{0.810} & \textit{1.00} \\  
FordB & 0.500 & \textbf{0.523} & \textit{0.500} & 0.500 & 0.500 & \textit{0.500} & \textit{0.94} & 0.500 & 0.94 & 0.500 & 0.01 \\  
BME & 0.611 & 0.687 & \textit{0.559} & 0.559 & 0.555 & \textbf{0.707} & \textit{1.21} & 0.707 & 1.21 & 0.504 & 0.83 \\  
\hline
Wins or ties over HC-DDTW & - & - &\multicolumn{3}{c}{\textbf{30/41}}   &\multicolumn{6}{c}{\textbf{31/41}}  \\  
Wins or ties over K-Shape & - & - &\multicolumn{3}{c}{\textbf{19/41}}   &\multicolumn{6}{c}{\textbf{21/41}}\\ 
Wins or ties over HC-L & - & - & \multicolumn{3}{c}{-}   &\multicolumn{6}{c}{\textbf{25/41}} \\
\bottomrule
\end{tabular}}
\end{table*} 

\subsection{Results}
We compare HC-AECS with important State-of-the-Art clustering mechanisms on time-series like HC with parametric Dynamic Time Warping (DTW)-derivative(DDTW) (HC-DDTW)\cite{luczak2016} and K-shape\cite{paparrizos2015k}. Authors of \cite{luczak2016} have already established that they perform better than K-means hence we don't include K-means in our analysis.

We have used RI as external clustering measure to perform unbiased comparison with the prior work as it has presented performance measure using RI. Further we have extended our analysis by using Normalized Mutual Information (NMI) \cite{strehl2002} as external clustering measure across a subset of considered 52 time-series. We have observed that NMI reflects also the same trend as obtained using RI as depicted in Table \ref{hcdtw_10}. As an example DistalPhalanxOAG, where NMI(HC-DDTW) = 0.414, NMI(K-Shape) = 0.252 and NMI(HC-AECS) = \{0.458, 0.321, 0.017\} for the distance measures CH, MA and ML respectively.  Here HC-AECS using CH gives the best performance as obtained using RI.


\begin{figure*}[ht]
\centering
\begin{subfigure}{.5\textwidth}
\centering
  \includegraphics[width=85mm,,height=42mm]{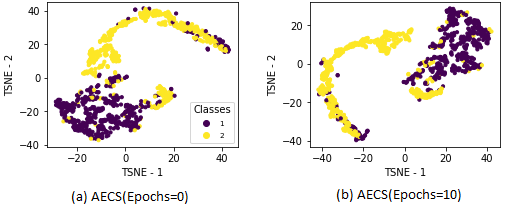}
  \label{fig:ecg}
\end{subfigure}%
\begin{subfigure}{.5\textwidth}  
\centering
  \includegraphics[width=95mm,height=40mm]{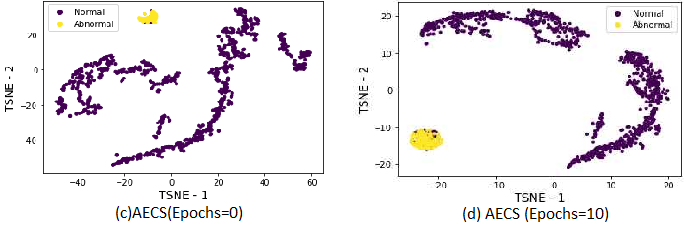}
  \label{fig:wafer}
\end{subfigure}
\caption{Visualisation of AECS using t-SNE for ECGFiveDays(univariate) (a) \& (b) and Wafer(multivariate) (c) \& (d)}
\label{fig3}
\end{figure*}

\subsubsection{Analysis on univariate time-series: Comparison w.r.t State-of-the-Art methods}

Table \ref{hcdtw_10} depicts comparison of clustering performance of HC-AECS using the proposed combination of distance measures CH, MA and ML with established State-of-the-Art clustering algorithms. Here, RI is used for comparative analysis. We also present the choice of best clustering based on $\mathcal{T}$ and corresponding distance measure following our proposed method. Additionally we have also demonstrated the performance of HC applied on raw time-series (HC-L) using the proposed distance measures. We indicate highest RI corresponding to the best distance measure in italics for HC-L and HC-AECS. Best RI achieved among all algorithms has been marked in bold. In summary, 76\% cases HC-AECS outperforms HC-DDTW the benchmark hierarchical clustering method and remains within the range of 0.001 to 0.281 with average variation of 0.065, and also 51\% cases outperforms K-Shape and remains within the range of 0.005 to 0.229 with average variation 0.1. 

We have also compared performance of our algorithm with existing benchmark classification results using 14 representative univarate time-series in Table \ref{classification}. The benchmark results for a dataset are reported for the algorithm which performs best among the 38 algorithms compared in \cite{bagnall2017great}. In this case we have applied HC-AECS on the given test data as benchmark accuracy are reported for test data only. Here, also HC-AECS generates very close classification benchmark results and outperforms multiple cases. In the cases where classification results are higher, HC-AECS only vary 0.12 in average (approximately 14\%) from benchmark classification results. The cases where test data contains less number of samples like DistalPhalanxOAG (139 test samples) and Lightning7 (73 test samples), HC-AECS lags behind.  

Figure \ref{fig3} depicts two-dimensional representation of proposed AECS using multi-layer undercomplete Seq2Seq auto-encoder. We use t-SNE\cite{maaten2008visualizing} with parameters: perplexity =30, learning rate=200 and iterations=100 to obtain 2-D plot for visualization. Two representative time-series ECGFivedays (univariate) and Wafer(UCI) (multivariate) are considered here. The clear separation of classes can be visualised using our learned compact representation.

\begin{table}[H]
\centering
\caption{Comparison of Rand Index (RI) of HC-AECS with benchmark classification accuracy}
\label{classification}
\resizebox{0.90\columnwidth}{!}{\begin{tabular}{lrrc}
\toprule
\textbf{Dataset} & \textbf{HC-AECS} & \multicolumn{2}{c}{\textbf{Benchmark \cite{bagnall2017great}}} \\ \cline{3-4}
& \textbf(RI) & \textbf{Accuracy} & \textbf{Algo} \\
\midrule 
DistalPhalanxOAG &  0.657 & 0.826 & HIVE-COTE\\ 
DistalPhalanxTW &  \textbf{0.823} & 0.698 & HIVE-COTE\\ 
MiddlePhalanxTW & \textbf{0.793} & 0.589 & SVML\\ 
ProximalPhalanxTW & 0.802 & 0.816 & RandF\\ 
SyntheticControl & 0.839 & 0.999 & HIVE-COTE \\ 
CricketX  & 0.720 & 0.830 & HIVE-COTE\\ 
CricketY &  0.762 & 0.837 & HIVE-COTE\\ 
CricketZ & 0.762 &	0.848 & HIVE-COTE \\ 
ECG5000 & 0.795 & 0.947 & HIVE-COTE \\ 
Lightning7 & 0.641 &  0.811 & HIVE-COTE\\ 
Adiac  & \textbf{0.937} & 0.815 & HIVE-COTE\\
SwedishLeaf & 0.848 & 0.968 & HIVE-COTE \\
Wafer & 0.795 & 1.000 & ST \\
Plane & 0.775 & 1.000 & HIVE-COTE \\
\bottomrule
\end{tabular}}
\end{table}

\begin{figure*}[htbp]
\centering
\begin{subfigure}{.3\textwidth}
\centering
  \includegraphics[width=45mm]{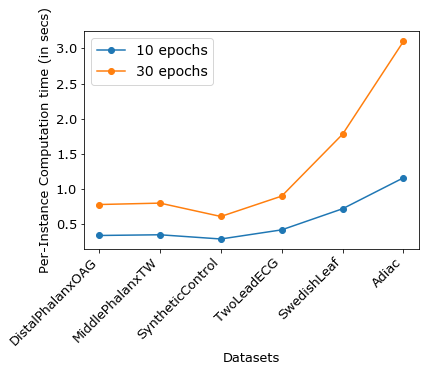}
  \caption{Comparison of per-instance computation time for HC-AECS on 10 and 30 epochs}
  \label{fig:sub1}
\end{subfigure}%
\begin{subfigure}{.7\textwidth}  
\centering
  \includegraphics[width=112mm,height=35mm]{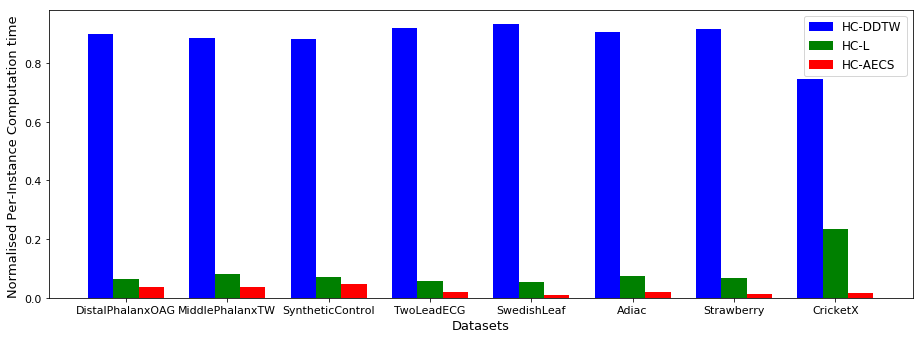}
  \caption{ Per-instance time for HC-DDTW, HC-L and HC-AECS on 8 time-series}
  \label{fig:sub2}
\end{subfigure}
\caption{Comparison of computation time for HC-AECS with benchmark algorithms}
\label{fig:test}
\end{figure*}

\subsubsection{Analysis on multivariate time-series}
We perform HC-AECS and measure RI on 9 representative out of 30 multivariate time-series from UCR and compare our results with benchmark classification algorithms like 1-NN using Euclidean distance(1NN-ED), dimension-indepedent DTW(1NN-DTW$_I$) and dimension-dependent DTW(1NN-DTW$_D$) \cite{bagnall2018,bagnall2017great} as presented in Table \ref{multi1}. We can see in 5 out of 9 datasets, our algorithms performs better than the benchmark algorithms. Additionally we compare our model with State-of-the-Art multivariate time-series classification algorithm MLSTM-FCN\cite{karim2019} using 2 datasets from UCI repository presented in Table \ref{multi2}. Our approach outperforms MLSTM-FCN by a significant margin for both the cases.

\begin{table}[htb]
\centering
\caption{Comparison of HC-AECS with Benchmark classification algorithms on UCR Multivariate time-series }
\label{multi1}
\resizebox{0.85\columnwidth}{!}{\begin{tabular}{lrrrr}
\toprule
\textbf{Dataset} &  \multicolumn{3}{c}{\textbf{1-NN}} & \textbf{HC-AECS} \\ \cline{2-4}
& \textbf{ED} & \textbf{DTW$_I$} & \textbf{DTW$_D$} &  \\
\midrule
AtrialFibrillation	& 	0.267 &	0.267 &	0.267 & \textbf{0.517} \\ 
ERing	&	0.133 &	0.133 &	0.133 & \textbf{0.709} \\ 
FingerMovement & \textbf{0.550} & 0.520 & 0.530 & 0.500 \\ 
HandMD &	0.278 &	0.306 &	0.231 & \textbf{0.539} \\ 
Handwriting	&	0.2 &	0.316 &	0.286 & \textbf{0.890} \\ 
Libras	&	0.833 &	\textbf{0.894} &	0.87 & 0.863 \\ 
NATOPS & 0.85 &	0.85 &	\textbf{0.883} & 0.723 \\ 
SelfRegulationSCP2 & 0.483 & 0.533 & \textbf{0.539} & 0.500 \\
StandWalkJump & 0.2 & 0.333 & 0.2 & \textbf{0.493} \\
\bottomrule
\end{tabular}}
\end{table}

\begin{table}[htb]
\centering
\caption{Comparison of HC-AECS with Benchmark classification algorithms on UCI Multivariate time-series }
\label{multi2}
\resizebox{0.85\columnwidth}{!}{\begin{tabular}{lrrr}
\toprule
\textbf{Dataset} &  \textbf{MLSTM-FCN} & \textbf{I-NN-DTW} & \textbf{HC-AECS} \\
\midrule
Wafer  & 0.906 &  0.671 & \textbf{0.929} \\
GesturePhase & 0.419 & 0.409 & \textbf{0.625} \\
\bottomrule
\end{tabular}}
\end{table}

\subsubsection{Computational time for HC-AECS}
HC-AECS consumes much less computation time as compared to HC-DDTW and HC with raw time-series due to the much reduced length of the latent representation. HC-DDTW requires a lot of computation power as performing DTW between two time-series of sequence length $n$ takes O($n^2$) time. So time taken for finding DTW between every pair of time-series reaches quadratic time complexity. On the other hand, all the distance measures used in our algorithm consumes much lower time than DTW to find dissimilarity between two time-series. Figure \ref{fig:sub2} presents a comparison of computing time between HC-DDTW, HC-L and HC-AECS, all of which exploits hierarchical clustering. Here per instance cluster time is computed for the algorithms which is total time taken divided by number of instances in the dataset. The time required by HC-AECS can be divided into three parts: (a) time required to learn AECS:($t_{aecs}$), (b) time required for clustering($t_c$),(c) time taken to choose best distance metric($t_v$). Figure \ref{fig:sub1} depicts computing time difference between epoch 10 and epoch 30. 
From our results, we conclude the computation time required for HC-AECS is approximately 27 times less than HC-DDTW. The time difference becomes more apparent in comparatively larger datasets like Adiac where HC-AECS performs almost 47 times faster than HC-DDTW. At the same time, $t_{aecs}$ does not vary significantly  with increase in epochs.


HC-AECS approach performs approximately 2.4 times faster than K-Shape in average considering the clustering time($t_c$). Furthermore in datasets like Coffee and Beef where number of timesteps is high and number of instances is relatively low, our approach performs significantly faster than K-Shape. For Coffee (Samples =56 , timesteps = 286) HC-AECS runs 16 times faster while for Beef (Samples = 60, timesteps = 470) its performs 18 times faster. As K-Shape is not applicable for multivariate time-series we perform the time comparison analysis only on univariate time-series.  

\section{Conclusion}
In this paper, we have presented a robust hierarchical clustering approach HC-AECS using a combination of hierarchical clustering with multi-layer under complete Seq2Seq auto-encoder representation. We have exploited compact latent representation of both univariate and multivariate time-series by selecting an appropriate distance measure. 
Proposed reduced representation AECS (Auto-Encoded Compact Sequence) and proper application of distance measures CH, MA and ML address high computational cost issue of hierarchical clustering along with capturing important representation. 

To this end, we have performed extensive analysis considering different distance measures and proposed a novel mechanism to choose the distance measure corresponding to the best clusters. We have used an internal clustering performance measure, Modified Hubert statistic $\mathcal{T}$ for best clustering selection. 

Our experimental analysis on UCR time-series from different application domains indicates the robust performance of our proposed scheme which is completely unsupervised. It outperforms $76\%$ w.r.t State-of-the-Art hierarchical clustering based time-series clustering method. Additionally our proposed method HC-AECS is applicable both on univariate and multivariate time-series. Presence of longer sequence length time-series are frequent in  healthcare, machine and manufacturing and smart-city application domains, here our method perfectly suits by representing a compact sequence which is constant across diverse length of time-series.

\bibliographystyle{named}
\bibliography{ijcai20}

\end{document}